\documentclass[sigconf]{acmart}

 \renewcommand\footnotetextcopyrightpermission[1]{} 
\usepackage{booktabs} 
\usepackage{url}
\usepackage{blindtext, graphicx}
\usepackage{float}
\usepackage{wrapfig}
\usepackage{array}
\usepackage{tikz}
\usetikzlibrary{positioning}
\usetikzlibrary{decorations.pathreplacing}

\usepackage{graphicx}
\usepackage{caption}
\usepackage{amsmath}
\usepackage{placeins}
\usepackage{multicol}
\newfloat{Eq.}{htbp}{loa}[equation]
\DeclareCaptionLabelFormat{nocount}{Eq.}

\begin{document}

\title{Correlated Time Series Forecasting using Deep Neural Networks: A Summary of Results}
\author{Razvan-Gabriel Cirstea, Darius-Valer Micu, Gabriel-Marcel Muresan, Chenjuan Guo, and Bin Yang\\
Department of Computer Science, Aalborg University, Denmark\\
\{rcirst16, dmicu16, gmures16\}@student.aau.dk\hspace{20pt}\{cguo, byang\}@cs.aau.dk}

\begin{abstract}
Cyber-physical systems often consist of entities that interact with each other over time. Meanwhile, as part of the continued digitization of industrial processes, various sensor technologies are deployed that enable us to record time-varying attributes (a.k.a., time series) of such entities, thus producing correlated time series. To enable accurate forecasting on such correlated time series, this paper proposes two models that combine convolutional neural networks (CNNs) and recurrent neural networks (RNNs). The first model employs a CNN on each individual time series, combines the convoluted features, and then applies an RNN on top of the convoluted features in the end to enable forecasting. The second model adds additional auto-encoders into the individual CNNs, making the second model a multi-task learning model, which provides accurate and robust forecasting. Experiments on two real-world correlated time series data set suggest that the proposed two models are effective and outperform baselines in most settings. 

This report extends the paper "Correlated Time Series Forecasting using Multi-Task Deep Neural Networks," to appear in ACM CIKM 2018, by providing additional experimental results. 
\end{abstract}

\keywords{Correlated time series; Deep learning; Multi-Task Learning}

\maketitle



\section{Introduction}

Complex cyber-physical systems (CPSs) often consist of multiple entities that interact with each other. 
%
With the continued digitization, various sensor technologies are deployed to record time-varying attributes of such entities, thus producing \emph{correlated time series}. 

For example, in an urban sewage system, sensors are deployed to capture time-varying concentration levels of different chemicals (e.g., NO$_3$ and NH$_4$) in sewage treatment plants. Different chemicals affect each other due to biological and chemical processes, thus making the different chemical time series correlated. 
As another example, in a vehicular transportation system~\cite{DBLP:journals/sigmod/GuoJ014}, traffic sensors (e.g., loop detectors and Bluetooth) are able to capture time-varying~\cite{DBLP:journals/geoinformatica/HuYJM17} traffic information (e.g., in the form of average speeds) of different road segments, which produces traffic time series~\cite{DBLP:journals/vldb/HuYGJ18}. 
Since the traffic on a road segment affects the traffic on other 
road segments, traffic time series on different road segments correlate with each other~\cite{,DBLP:journals/vldb/YangDGJH18,DBLP:journals/pvldb/DaiYGJH16}. 

Accurate forecasting of correlated time series have the potential to reveal holistic system dynamics of the underlying CPSs, including identifying trends, predicting future behavior~\cite{DBLP:journals/pvldb/0002GJ13}, and detecting anomalies~\cite{DBLP:conf/mdm/Kieu0J18}, which are important to enable effective operations of the CPSs. For example, in an sewage system, time series forecasting enables identifying the changing trends of different chemicals, early warning of high concentrations of toxic chemicals, and predicting the effect of incidents, e.g., drought or rains, which enables more effective and purposeful operations of the sewage system. Similarly, in an intelligent transportation system, analyzing traffic time series enables travel time forecasting, early warning of congestion, and predicting the effect of incidents, which benefit drivers and fleet owners. 

To enable accurate and robust correlated time series forecasting, we propose two novel non-linear forecasting algorithms based on deep neural networks---a Convolutional Recurrent Neural Network (CRNN) and an Auto-Encoder Convolutional Recurrent Neural Network (AECRNN). 
In CRNNs, we first consider each of the correlated time series independently and feed each time series into a 1-dimensional convolutional neural network (CNN). The usage of the CNNs helps us learn features for each individual time series. Next, the convoluted time series features are merged together, which is then fed into a Recurrent Neural Network (RNN), with the aim of learning the sequential information while considering the \emph{correlations} among different time series. 

In AECRNNs, we add additional auto-encoders into CRNN. The output of convoluted time series are not only merged together to be fed into the RNN. In addition, each convoluted time series is also reconstructed back to the original time series. Then, the objective function combines the error of the prediction by the RNN and the reconstruction discrepancy by the auto-encoders, making AECRNN a multi-task learning model. The use of auto-encoders makes the CNNs also learn representative features of each time series, but not only distinct features for predicting future values. 
In other words, the auto-encoders work as an extra regularization which avoids overfitting, disregards outliers, and thus provides more robust forecasting. 

To the best of our knowledge, this is the first study that combines CNNs and RNNs in a unified framework with the help of multi-task learning to enable accurate forecasting for correlated time series. Experiments on a large real-word chemical concentration time series from a sewage treatment plant and a time series data set from Google Trends offer evidence that the proposed methods are accurate and robust.

\section{Preliminaries}\label{preliminaries}


\par A \textit{time series} $X^{(i)} = \langle x^{(i)}_1, x^{(i)}_2, \ldots, x^{(i)}_m \rangle$ is a time-ordered sequence of measurements, where measurement $x^{(i)}_k$ is recorded at time stamp $t_k$ and we have $t_j < t_k$ if $1 \leq j<k\leq m$.
Usually, the time interval between two consecutive measurements is constant, i.e., $t_{j+1}-t_{j}=t_{k+1}-t_{k}, 1\leq j, k<m$. 
%
%

\begin{center}
 \begin{tabular}{|c|c|} 
 \hline
 Symbol & Meaning 
 \\\hline
 $X$ & $X = \{ X^{(1)}, X^{(2)}, ..., X^{(n)} \}$\\
 & A set of correlated time series \\\hline
 
 $X^{(i)}$ & $X^{(i)} = \langle x_1^{(i)}, x_2^{(i)}, ... , x_m^{(i)} \rangle$, $1 \leq i \leq n$ \\
 & A sequence of measurements from time series $X^{(i)}$ \\\hline
 
 $\hat{X}^{(i)}$ & $\hat{X}^{(i)} = \langle \hat{x}_1^{(i)}, \hat{x}_2^{(i)}, ... , \hat{x}_m^{(i)} \rangle$, $1 \leq i \leq n$ \\  
 & Reconstruction of time series $X^{(i)}$ \\\hline
 
 $Z$ & $Z = \langle z_1, z_2, ... , z_p \rangle$ \\ 
 & A sequence of predicted values \\\hline
 
 $x_{t}^{(i)}$ & The measurement at time $t$ of $X^{(i)}$, $1 \leq t \leq m$ \\
 & \\\hline
 
 $\hat{x}_{t}^{(i)}$ & The measurment at time $t$ of $\hat{X}^{(i)}$, $1\leq t \leq m$ \\
 & \\\hline
 
 $z_{t}^{(i)}$ & Prediction at time $t$ of $Z$, $1\leq t \leq m$ \\
 & \\\hline
 
 
 
 $n$ & $n = |X|$\\
 & Number of correlated time series \\\hline
 
 $l$ & Length of a model input \\\hline

 $p$ & $p = |Z|$\\
 & Prediction size \\\hline
\end{tabular}
\captionof{table}{Notation}
\label{general_notations}
\end{center}

\par A \textit{correlated time series set} is denoted as $X=\langle X^{(1)}, X^{(2)}, \ldots, X^{(n)} \rangle$, where time series in $X$ are correlated with each other. 
For example, in the sewage treatment example, we have $X=\langle X^{(1)}, X^{(2)}, X^{(3)} \rangle$, where $X^{(1)}$, $X^{(2)}$, and $X^{(3)}$ represent the time series of NH$_4$, NO$_3$, and O$_2$, respectively.  
Important notation is shown in Table~\ref{general_notations}. 

\noindent
\textbf{Problem statement: } 
Given a correlated time series set $X = \langle X^{(1)}$, $X^{(2)}$, $\ldots$, $X^{(n)} \rangle$, 
we aim at predicting the future measurements of a specific, target time series in $X$. Without loss of generality, the first time series $X^{(1)}$ is chosen as the target time series. 
More specifically, we assume that the given time series in $X$ have measurements covering a window $[t_{a+1}, t_{a+l}]$ that contains $l$  time stamps, and we aim at predicting time series
$X^{(1)}$'s measurements in a future window $[t_{a+l+1}, t_{a+l+p}]$. We call this problem \textit{$p$-step ahead forecasting}. 


\section{Related Work}\label{related_work}


We summarize related studies on time series forecasting in Table~\ref{related_work_summary}. We consider two dimensions---
linear vs. non-linear forecasting and single vs. multiple time series.

\begin{table}[!htp]
\centering \small
\begin{tabular}{|l|l|l|}
\hline
              & Single                                                           & Multiple                                                                \\ \hline
Linear        & \cite{ExponentialMA,arima}                                           &\cite{multiple_regression,DBLP:journals/pvldb/0002GJ13,DBLP:conf/sigmod/MatsubaraSF14}                      \\ \hline
Non-Linear & NN, \cite{NNVsArima, RNNAndARIMA, ARIMA_RNN,UberPaper,ASSAAD200841} & RNN, LSTM, \cite{Pang:2017:RAM:3155133.3155172,Ecoweb,CNN_LSTM_RAIN} \\
 & &  \textbf{CRNN, AECRNN} \\ \hline
\end{tabular}
\caption{Related Work}
\label{related_work_summary}
\end{table}

We first consider linear methods for single time series. 
Here, methods such as Exponentially Weighted Moving Average (EWMA) \cite{ExponentialMA} and Autoregressive Integrated Moving Average (ARIMA) \cite{arima} are simple yet effective for modeling linear time series and nowadays are commonly used as baseline methods. 
%

Linear methods for multiple time series also exist, e.g., multiple linear regression~\cite{multiple_regression}, spatio-temporal hidden Markov models~\cite{DBLP:journals/pvldb/0002GJ13}, multi-level chain model~\cite{DBLP:conf/sigmod/MatsubaraSF14}. 

Neural networks (NN) are able to model non-linear relationships, which are often employed to enable non-linear forecasting models. For example, Recurrent Neural Networks (RNN) and Long Short Term Memory (LSTM)~\cite{LSTMFirst} are able to provide non-linear time series forecasting.   
Hybrid models that combine ARIMA with RNN are also proposed~\cite{RNNAndARIMA,ARIMA_RNN}.  
%
The hybrid models start with ARIMA to identify linear dependencies and the resulting residuals are fed to an RNN which captures the nonlinear dynamics.  
A boosting method using RNNs is also proposed~\cite{ASSAAD200841}. 
In another paper~\cite{UberPaper}, a novel model is proposed as an extension to a classical LSTM. The model first builds an LSTM auto-encoder, which extracts automatically features from the bottleneck layer. The extracted features are merged with the original input data, which are fed together to another LSTM for predictions. 
%

In this paper, we aim at providing non-linear forecasting models that are able to support multiple time series, i.e., the \textit{right bottom cell} of Table~\ref{related_work_summary}. 
%
\textit{Pang et al.} propose multivariate time series convolutional neural network (MTCNN)~\cite{Pang:2017:RAM:3155133.3155172} that uses a CNN to extract features from a multivariate time series before passing the results to a fully connected neural network layer, which provides better forecasting accuracy compared to regular CNNs.  
We propose two models, namely CRNN and AECRNN, which outperforms existing methods in most settings. Similar to MTCNN, we first use CNNs to extract features from multipe time series. However, we instead use RNN in the proposed CRNN and AECRNN. In addition, AECRNN also incorporates auto-encoders, which enables effective and robust forecasting. This also makes AECRNN a multi-task learning model. 
Another similar approach is ConvLSTM~\cite{CNN_LSTM_RAIN}, which combines CNN with LSTM to enable forecasting of a series of 2D radar maps. 

Non-deep learning methods also exist. For example, a non-linear forecasting algorithm using Lotka-Volterra equation from biology is proposed~\cite{Ecoweb}. It assumes that multiple time series are competing each other for a finite resource, e.g., predators competing for food. This assumption limits the scope where the method can be applied, and thus we do not compare with it in the experiments.    
%

Similar to AECRNN, T2INet~\cite{tung} is also a multi-task learning model. It employs CNN and auto-encoders to enable classification and clustering, but not for prediction.

\begin{figure*}[!htb]
    \centering
    \captionsetup{justification=centering}
    \includegraphics[width=\textwidth]{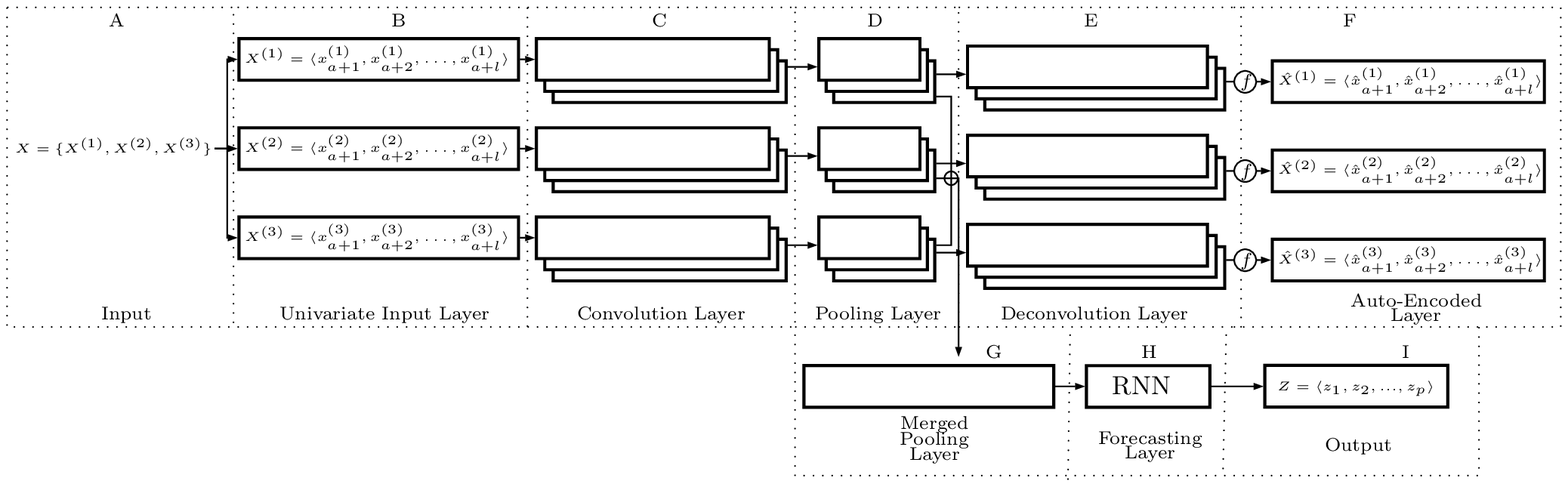}
    \caption{Auto-Encoder Convolutional Recurrent Neural Network (AECRNN)}
    \label{fig:MAECNNRNN}
\end{figure*}

\begin{figure*}[!htb]
    \centering
    \captionsetup{justification=centering}
    \includegraphics[width=\textwidth]{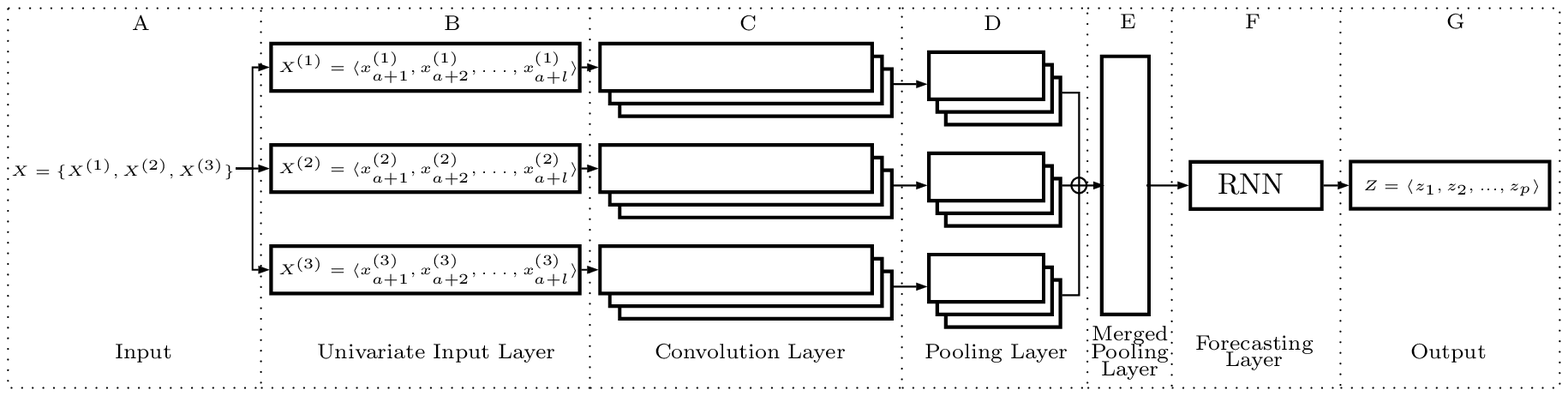}
    \caption{Convolutional Recurrent Neural Network (CRNN)}
    \label{fig:MCRNN}
\end{figure*}

\section{Proposed Models}\label{proposed_model}

\subsection{Convolutional Recurrent Neural Network (CRNN)}
\label{sec:crnn}


We propose a CRNN that utilizes a combination of Convolutional Neural Networks (CNNs) and Recurrent Neural Networks (RNNs) to enable p-step ahead forecasting for a set of correlated time series. 
%
%
CNNs are used successfully for classifying images by learning the features and patterns of the images. RNNs are able to capture dependencies of a sequence of values, thus being good at forecasting future values. 
%
This motivates us to first use CNNs to extract distinctive features of each of the correlated multiple time series, and then to apply RNNs on top of the combined output of CNNs to predict near future values of the target time series. We call this model CRNN, which is presented in Fig.~\ref{fig:MCRNN}.

Specifically, CRNN takes as input multiple time series, where each input time series contains $l$ measurements. Part A of Fig~\ref{fig:MCRNN} shows an example where $|X|=3$ correlated time series $X^{(1)}, X^{(2)}$, and $X^{(3)}$ are fed into CRNN as input. 
CRNN outputs a single time series, say $Z=\langle z_1, z_2, \ldots, z_p \rangle$, that contains $p$ measurements which are the predicted $p$ measurements of the target time series $X^{(1)}$ in the near future (see Part G of Fig~\ref{fig:MCRNN}).  

In the CRNN, we first treat each time series in $X$ independently. 
In particular, we treat each time series as a $1\times l$ matrix, as shown in part B of Fig. \ref{fig:MCRNN}.  
Next, we apply convolutions on each time series at the convolutional layer (see part C of Fig.~\ref{fig:MCRNN}). In particular, we apply $\alpha$, e.g., 3, in Fig.~\ref{fig:MCRNN}, filters to conduct convolutions, with the aim to extract distinctive features in the individual input time series. This produces $\alpha$ matrices with size of $1\times l$. 
Next, in the pooling layer (see part D of Fig.~\ref{fig:MCRNN}), for each matrix, we apply a max-pooling operator to capture the most representative features of the time series as a $1\times \frac{l}{2}$ matrix by using a 1$\times$ 2 window with a stride set to 2. Thus, we obtain a total of $\alpha$ matrices of size $1\times \frac{l}{2}$ for each input time series. 
Note that we may apply the convolution and pooling layers multiple times. 
After convolution and pooling, we have $|X|$ cubes of size $\alpha\times 1\times \frac{l}{2}$ as the output of the pooling layers (see part D Fig. \ref{fig:MCRNN}-D). 
So far, the work of CNNs finishes. 

Next, the $|X|$ cubes are concatenated into an $n$-dimensional vector (see part E of Fig. \ref{fig:MCRNN}), where $n=|X|\times\alpha\times 1\times \frac{l}{2}$ to be fed into an RNN (see part F of Fig. \ref{fig:MCRNN}) for the predicting purpose. We obtain $Z$ as the near future measurements of the target time series (see Part G of Fig~\ref{fig:MCRNN}). 
%
%
The objective function of CRNN is 
%
%
%
%
%
\vspace{-5pt}
\begin{equation}\label{eq:loss_function}
        J_1 = \frac{1}{p} \sum_{i=1}^{p}{\mathit{Error}(z_{i},{x}^{(1)}_{a+l+i})}
\vspace{-5pt}
\end{equation}
%
%
Here, $\mathit{Error}(\cdot, \cdot)$ is an error function that measures the discrepancy (e.g., mean square error) between the predicted measurement $z_{i}$ and the ground truth measurement ${x}^{(1)}_{a+l+i}$ at time stamp $a+l+i$. 

Note that the RNN can be easily extended to enable forecasting future measurements of all time series, but not just a single target time series.
%




\subsection{Auto Encoder CRNN (AECRNN)}
\label{sec:aecrnn}

In AECRNN, we incorporate an auto-encoder to each CNN (see Fig~\ref{fig:MAECNNRNN}). 
The intuition behind AECRNN is to use auto-encoders to learn robust features and ignore features that represent outliers. The auto-encoders also work as an additional regularization to enable learning most representative features for all input time series but not overfit to features that are specific to forecasting the target time series in the training data. 

After the pooling layer, we not only concatenate the output cubes of the pooling layer, but also feed the output cubes to an additional deconvolution layer (see Part E of Fig~\ref{fig:MAECNNRNN}). 
In particular, each cube is deconvoluted into $\alpha$ matrices with the same size as the matrices in part C. Then, we obtain $|X|$ groups of matrices, where each group has $\alpha$ matrices.

Next, we apply Sigmoid activation function to each matrix group to produce a $1\times l$ matrix. This corresponds $|X|$ reconstructed time series, e.g., $\hat{X}^{(1)}$, $\hat{X}^{(2)}$, and $\hat{X}^{(3)}$, where $\hat{X}^{(i)}=\langle\hat{x}_{a+1}^{(i)}, \hat{x}_{a+2}^{(i)}, \ldots, \hat{x}_{a+l}^{(i)}\rangle$. 

The objective function of the additional auto encoders is 
%
$$J_2 = \frac{1}{l} \cdot \frac{1}{|X|} \cdot \sum_{k=1}^{|X|} \sum_{i=1}^{l} {\textit{Error}(\hat{x}^{(k)}_{a+i}, {x}^{(k)}_{a+i})}.$$
%
It measures the discrepancy between the reconstructed measurement $\hat{x}^{(k)}_{a+i}$ and the original, ground-truth measurement ${x}^{(k)}_{a+i}$ at the $(a+i)$-th time stamp over all $|X|$ time series. 

The final objective function of AECRNN is $J=J_1 + J_2$. This makes AECRNN a multi-task learning model, where one task is to forecast the $p$ future measurements of the target time series (i.e., $J_1$) and the other tasks are to reconstruct $|X|$ time series' $l$ known measurements (i.e., $J_2$). 

%



%
\section{Empirical Study}\label{testing_scenarious}

\subsection{Experimental Setup}

\noindent
\textbf{Data Sets: }We use two time series data sets in the experiments. 
%
The first data set was provided by a sewage treatment center from Aalborg, Denmark. 
The sewage treatment center has 6 tanks and three different sensors are deployed in each tank to measure the concentrations of three different chemicals, NH$_4$, NO$_3$, and O$_2$, every 2 minutes. The data covers a period of 3 years in total.  

\begin{figure}[!ht]
    \centering
    \captionsetup{justification=centering}
    \includegraphics[width=\columnwidth]{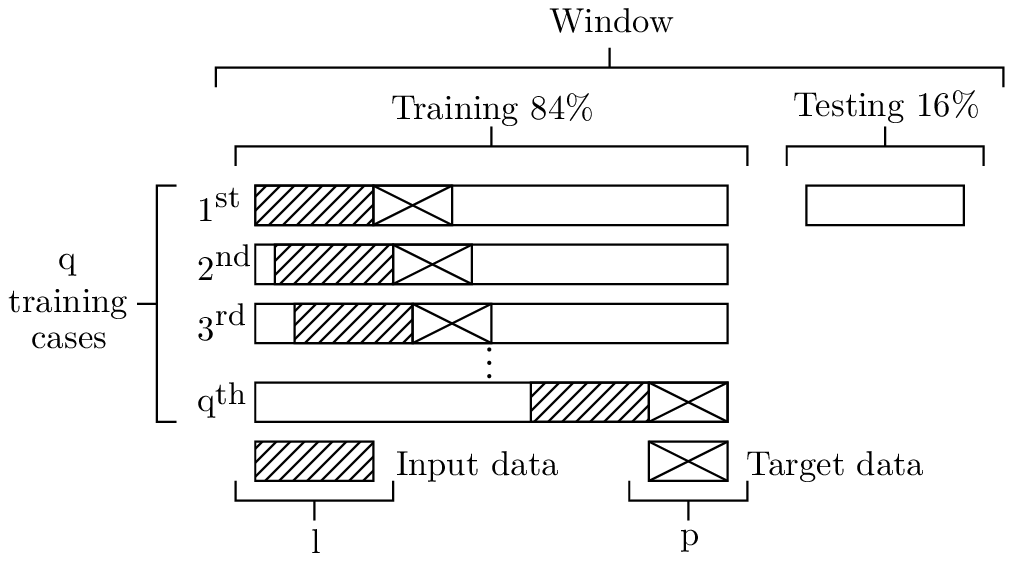}
    \caption{Training vs Testing}
    \label{fig:LearnProcess}
\end{figure}

%
We consider the three time series of the three chemicals from a specific tank as correlated time series. We choose multiple windows over the 3-year period to test the proposed methods. For each window, we use the first 84\% of the data for training and validation, and the remaining 16\% of data for testing.
When learning the proposed models, we further segment the training data into multiple training cases using a sliding window. 
In particular, in each segment, we use a sequence of $l$ measurements as the \emph{input data} to the proposed models and use the immediately following $p$ measurements as the \emph{ground truth} target data to calculate the predication errors (i.e., $J_1$) to enable back-propagation.
Fig.~\ref{fig:LearnProcess} illustrates the process. 
%
%
Finally, evaluation of the learned models was conducted on the reserved 16\% testing data. 

\par Validation and hyper-parameters selection on the sewage data set for all algorithms was performed on tank 4, window 1. This data set was excluded from all the results presented later in the paper. The reason for the exclusion is to eliminate any conflict generated by the intersection of validation data with testing data. 

%

\par
The second data set was taken from Google Trends. A time series represents a keyword's monthly search popularity over the last 14 years. 
For this data set, we have selected keywords \textit{potatoes} and \textit{brown sugar} due to their high correlation.
%


\noindent
\textbf{Parameters: }
We vary three \emph{problem parameters} in the experiments. Specifically, we vary the number of time series, i.e., the cardinality of the correlated time series set $|X|$, from 1, 2, to 3. When $|X|=1$, we only consider a single time series, which is also the target time series. Then, we add 1 and 2 additional correlated time series with the target time series, respectively. Next, we vary the length $l$ of time series when training. We also vary $p$ to enable $p$-step ahead forecasting, where a large $p$ means forecasting far into the future.

%



%
In addition, we vary \emph{solution parameters (a.k.a. hyper-parameters)} that are specific to solutions. In particular, for the proposed deep learning based methods, we vary the number of convolution and pooling layers (1, 2, and 3), number of filters per convolution layer (2, 3, 4, 5, 8, 10, and 16), the filer size (1, 2, 3, 5, and 10), and hidden state size in RNNs (3, 4, 5, 6). We identify the optimal solution parameters for each problem parameter setting.



\noindent
\textbf{Baselines: } We consider 5 baselines. \emph{Yesterday} is a simple, linear method that propagates the last known value of the time series for the entire prediction window, which is commonly used in financial time series prediction. Methods \emph{ARIMA}, \emph{RNN}, \emph{LSTM}, and \emph{MTCNN} are covered in Section~\ref{related_work}. 

\noindent
\textbf{Implementation Details: } All methods are implemented in Python 3.6, where the deep learning methods are implemented using Tensorflow 1.7. A computer with Intel i7-4700MQ CPU, 4 cores, 16 GB RAM is used to conduct all experiments. 

\subsection{Experimental Results}
We use both root mean square error (RMSE) and mean absolute percentage error (MAPE) to evaluate accuracy.

\subsubsection{Results on Sewage Data}
Tables~\ref{table:rmseResult} and \ref{table:rmseResultTable}
show the average RMSE values with standard deviations while varying problem parameters $|X|$, $l$, and $p$. 
Tables~\ref{table:mapeResult} and \ref{table:mapeResultTable} 
%
show the average MAPE values with standard deviations while varying problem parameters $|X|$, $l$, and $p$. \\

\begin{figure*}[!h]
\begin{tabular}{llllllll}
\hline
 |X|, l, p   & YESTERDAY            & ARIMA                         & RNN                           & LSTM                          & MTCNN                & CRNN                         & AECRNN                     \\
\hline
 1\_200\_100  & \text{1.31$\pm$0.32} & \text{1.49$\pm$0.26}          & \textbf{\text{0.80$\pm$0.06}} & \text{0.86$\pm$0.06}          & \text{0.95$\pm$0.16} & \text{0.93$\pm$0.31}          & \text{1.00$\pm$0.38}          \\
 2\_200\_100  & \text{1.31$\pm$0.32} & \text{1.49$\pm$0.26}          & \text{0.75$\pm$0.12}          & \text{0.86$\pm$0.10}          & \text{0.96$\pm$0.16} & \text{0.68$\pm$0.11}          & \textbf{\text{0.61$\pm$0.07}} \\
 3\_200\_100  & \text{1.31$\pm$0.32} & \text{1.49$\pm$0.26}          & \text{0.81$\pm$0.27}          & \text{0.93$\pm$0.21}          & \text{0.87$\pm$0.09} & \textbf{\text{0.77$\pm$0.29}} & \text{0.82$\pm$0.23}          \\ \hline
 2\_10\_100   & \text{1.24$\pm$0.33} & \text{1.49$\pm$0.52}          & \text{0.78$\pm$0.37}          & \textbf{\text{0.73$\pm$0.30}} & \text{0.82$\pm$0.35} & \text{0.79$\pm$0.35}          & \text{0.80$\pm$0.36}          \\
 2\_50\_100   & \text{1.18$\pm$0.21} & \text{1.58$\pm$0.39}          & \text{0.84$\pm$0.22}          & \text{0.92$\pm$0.30}          & \text{1.00$\pm$0.41} & \text{0.96$\pm$0.34}          & \textbf{\text{0.84$\pm$0.37}} \\
 2\_100\_100  & \text{1.32$\pm$0.27} & \text{2.86$\pm$1.85}          & \text{1.07$\pm$0.30}          & \text{1.08$\pm$0.26}          & \text{1.12$\pm$0.39} & \text{1.09$\pm$0.36}          & \textbf{\text{0.91$\pm$0.29}} \\ \hline
 2\_200\_1    & \text{0.11$\pm$0.01} & \textbf{\text{0.07$\pm$0.01}} & \text{0.45$\pm$0.16}          & \text{0.40$\pm$0.10}          & \text{0.30$\pm$0.08} & \text{0.35$\pm$0.16}          & \text{0.70$\pm$0.59}          \\
 2\_200\_25   & \text{1.02$\pm$0.15} & \text{0.99$\pm$0.13}          & \text{0.78$\pm$0.23}          & \text{0.76$\pm$0.24}          & \text{1.04$\pm$0.31} & \text{0.62$\pm$0.30}          & \textbf{\text{0.58$\pm$0.24}} \\
 2\_200\_50   & \text{1.35$\pm$0.29} & \text{1.49$\pm$0.21}          & \text{0.74$\pm$0.26}          & \text{0.68$\pm$0.09}          & \text{0.76$\pm$0.16} & \text{0.81$\pm$0.30}          & \textbf{\text{0.63$\pm$0.21}} \\
\hline
\end{tabular}
\captionof{table}{RMSE, sewage data set, 1 tank. 
}
\vspace{-10pt}
 \label{table:rmseResult}
\end{figure*}
\begin{figure*}[!h]
\begin{tabular}{rccccccc}
\hline
 |X|, l, p   & YESTERDAY                                                    & ARIMA                                                          & RNN                                                         & LSTM                                                        & MTCNN                                                        & CRNN                                                                    & AECRNN                                                                        \\
\hline
2\_10\_100   & \text{\textcolor{white}{0}1.02$\pm$\textcolor{white}{0}0.36} & \text{\textcolor{white}{0}1.42$\pm$\textcolor{white}{0}1.00}                & \text{\textcolor{white}{0}0.69$\pm$\textcolor{white}{0}0.30} & \text{\textcolor{white}{0}0.71$\pm$\textcolor{white}{0}0.28} & \text{\textcolor{white}{0}0.60$\pm$\textcolor{white}{0}0.31} & \text{\textcolor{white}{0}0.64$\pm$\textcolor{white}{0}0.33} & \text{\textbf{\text{\textcolor{white}{0}0.57$\pm$\textcolor{white}{0}0.30}}}  \\
 2\_50\_100   & \text{\textcolor{white}{0}1.00$\pm$\textcolor{white}{0}0.36} & \text{\textcolor{white}{0}1.15$\pm$\textcolor{white}{0}0.47}                & \text{\textcolor{white}{0}0.74$\pm$\textcolor{white}{0}0.28} & \text{\textcolor{white}{0}0.79$\pm$\textcolor{white}{0}0.29} & \text{\textcolor{white}{0}0.64$\pm$\textcolor{white}{0}0.35} & \text{\textcolor{white}{0}0.74$\pm$\textcolor{white}{0}0.35} & \text{\textbf{\text{\textcolor{white}{0}0.58$\pm$\textcolor{white}{0}0.31}}}  \\
 2\_100\_100  & \text{\textcolor{white}{0}1.08$\pm$\textcolor{white}{0}0.59} & \text{\textcolor{white}{0}1.18$\pm$\textcolor{white}{0}0.66}                & \text{\textcolor{white}{0}0.83$\pm$\textcolor{white}{0}0.37} & \text{\textcolor{white}{0}0.83$\pm$\textcolor{white}{0}0.38} & \text{\textcolor{white}{0}0.73$\pm$\textcolor{white}{0}0.44} & \text{\textcolor{white}{0}0.95$\pm$\textcolor{white}{0}0.50} & \text{\textbf{\text{\textcolor{white}{0}0.66$\pm$\textcolor{white}{0}0.38}}}  \\\hline
2\_50\_1     & \text{\textcolor{white}{0}0.09$\pm$\textcolor{white}{0}0.03} & \text{\textbf{\text{\textcolor{white}{0}0.05$\pm$\textcolor{white}{0}0.02}}}& \text{\textcolor{white}{0}0.21$\pm$\textcolor{white}{0}0.13} & \text{\textcolor{white}{0}0.15$\pm$\textcolor{white}{0}0.09} & \text{\textcolor{white}{0}0.14$\pm$\textcolor{white}{0}0.08} & \text{\textcolor{white}{0}0.13$\pm$\textcolor{white}{0}0.08} & \text{\textcolor{white}{0}0.20$\pm$\textcolor{white}{0}0.13}                  \\
 2\_50\_25    & \text{\textcolor{white}{0}0.93$\pm$\textcolor{white}{0}0.22} & \text{\textcolor{white}{0}0.74$\pm$\textcolor{white}{0}0.25}                & \text{\textcolor{white}{0}0.44$\pm$\textcolor{white}{0}0.22} & \text{\textcolor{white}{0}0.45$\pm$\textcolor{white}{0}0.24} & \text{\textcolor{white}{0}0.45$\pm$\textcolor{white}{0}0.25} & \text{\textcolor{white}{0}0.42$\pm$\textcolor{white}{0}0.26} & \text{\textbf{\text{\textcolor{white}{0}0.38$\pm$\textcolor{white}{0}0.21}}}  \\
 2\_50\_50    & \text{\textcolor{white}{0}1.03$\pm$\textcolor{white}{0}0.28} & \text{\textcolor{white}{0}1.03$\pm$\textcolor{white}{0}0.40}                & \text{\textcolor{white}{0}0.66$\pm$\textcolor{white}{0}0.30} & \text{\textcolor{white}{0}0.55$\pm$\textcolor{white}{0}0.31} & \text{\textcolor{white}{0}0.57$\pm$\textcolor{white}{0}0.34} & \text{\textcolor{white}{0}0.54$\pm$\textcolor{white}{0}0.29} & \text{\textbf{\text{\textcolor{white}{0}0.47$\pm$\textcolor{white}{0}0.28}}}  \\
\hline
\end{tabular}
\captionof{table}{RMSE, sewage data set, 6 tanks. 
}\vspace{-10pt}
 \label{table:rmseResultTable}
\end{figure*}
\vspace{-10pt}
\begin{figure*}[!ht]
\begin{tabular}{llllllll}
\hline
 |X|, l, p   & YESTERDAY             & ARIMA                         & RNN                           & LSTM                          & MTCNN                 & CRNN                         & AECRNN                      \\
\hline
 1\_200\_100  & \text{16.59$\pm$2.70} & \text{18.77$\pm$2.65}         & \textbf{\text{9.50$\pm$1.39}} & \text{10.16$\pm$2.60}         & \text{10.49$\pm$0.80} & \text{10.78$\pm$1.26}         & \text{11.33$\pm$2.21}          \\
 2\_200\_100  & \text{16.59$\pm$2.70} & \text{18.77$\pm$2.65}         & \text{9.43$\pm$1.45}          & \text{10.79$\pm$1.61}         & \text{11.08$\pm$0.91} & \textbf{\text{7.91$\pm$0.37}} & \text{8.45$\pm$0.61}           \\
 3\_200\_100  & \text{16.59$\pm$2.70} & \text{18.77$\pm$2.65}         & \text{9.59$\pm$1.85}          & \text{10.77$\pm$0.57}         & \text{10.24$\pm$1.45} & \textbf{\text{9.08$\pm$1.74}} & \text{9.59$\pm$0.93}           \\ \hline
 2\_10\_100   & \text{18.23$\pm$2.93} & \text{20.85$\pm$3.98}         & \text{9.86$\pm$2.77}          & \textbf{\text{9.12$\pm$1.93}} & \text{10.01$\pm$2.75} & \text{9.48$\pm$2.44}          & \text{9.55$\pm$2.56}           \\
 2\_50\_100   & \text{16.41$\pm$2.25} & \text{20.75$\pm$5.37}         & \text{10.77$\pm$3.10}         & \text{11.90$\pm$3.15}         & \text{12.20$\pm$3.66} & \text{11.83$\pm$3.30}         & \textbf{\text{9.56$\pm$2.52}}  \\
 2\_100\_100  & \text{17.90$\pm$1.58} & \text{36.09$\pm$24.37}        & \text{12.62$\pm$2.89}         & \text{12.92$\pm$2.39}         & \text{12.43$\pm$1.84} & \text{12.78$\pm$3.30}         & \textbf{\text{11.04$\pm$1.37}} \\ \hline
 2\_200\_1    & \text{1.42$\pm$0.25}  & \textbf{\text{0.78$\pm$0.05}} & \text{5.68$\pm$2.04}          & \text{4.72$\pm$1.07}          & \text{3.54$\pm$0.92}  & \text{3.97$\pm$1.40}          & \text{7.99$\pm$5.88}           \\
 2\_200\_25   & \text{12.95$\pm$3.35} & \text{11.67$\pm$2.61}         & \text{10.04$\pm$2.74}         & \text{9.98$\pm$3.20}          & \text{11.38$\pm$1.92} & \text{7.42$\pm$2.70}          & \textbf{\text{7.12$\pm$1.92}}  \\
 2\_200\_50   & \text{17.49$\pm$2.47} & \text{18.76$\pm$2.73}         & \text{9.42$\pm$2.36}          & \text{9.13$\pm$2.45}          & \text{8.77$\pm$0.83}  & \text{10.06$\pm$2.28}         & \textbf{\text{7.81$\pm$1.63}}  \\
\hline
\end{tabular}
\captionof{table}{MAPE, sewage data set, 1 tank}
 \label{table:mapeResult}\vspace{-10pt}
\end{figure*}
\begin{figure*}[!ht]
\begin{tabular}{rccccccc}
\hline
 |X|, l, p   & YESTERDAY                                                    & ARIMA                                                          & RNN                                                         & LSTM                                                        & MTCNN                                                        & CRNN                                                                    & AECRNN                                                                        \\
\hline
2\_10\_100   & \text{25.51$\pm$12.99}                                      & \text{26.57$\pm$15.11}                                                       & \text{19.17$\pm$12.01}                                       & \text{19.88$\pm$13.52}                                       & \text{14.46$\pm$\textcolor{white}{0}7.73}                    & \text{16.87$\pm$10.54}                                       & \text{\textbf{\text{14.22$\pm$\textcolor{white}{0}7.92}}}                     \\
 2\_50\_100   & \text{25.95$\pm$13.64}                                      & \text{26.50$\pm$16.75}                                                       & \text{22.93$\pm$17.10}                                       & \text{24.09$\pm$17.91}                                       & \text{15.65$\pm$\textcolor{white}{0}8.23}                    & \text{23.08$\pm$20.31}                                       & \text{\textbf{\text{15.28$\pm$10.09}}}                                        \\
 2\_100\_100  & \text{28.25$\pm$14.34}                                      & \text{29.81$\pm$17.22}                                                       & \text{25.61$\pm$19.25}                                       & \text{27.30$\pm$24.59}                                       & \text{19.78$\pm$12.22}                                       & \text{29.80$\pm$23.26}                                       & \text{\textbf{\text{18.32$\pm$11.82}}}                                        \\\hline
2\_50\_1     & \text{\textcolor{white}{0}2.21$\pm$\textcolor{white}{0}1.09}& \text{\textbf{\text{\textcolor{white}{0}0.98$\pm$\textcolor{white}{0}0.41}}} & \text{\textcolor{white}{0}4.76$\pm$\textcolor{white}{0}3.05} & \text{\textcolor{white}{0}3.63$\pm$\textcolor{white}{0}2.19} & \text{\textcolor{white}{0}3.34$\pm$\textcolor{white}{0}1.84} & \text{\textcolor{white}{0}3.00$\pm$\textcolor{white}{0}1.63} & \text{\textcolor{white}{0}5.04$\pm$\textcolor{white}{0}3.36}                  \\
 2\_50\_25    & \text{23.61$\pm$10.77}                                      & \text{16.95$\pm$10.95}                                                       & \text{11.15$\pm$\textcolor{white}{0}6.09}                    & \text{11.81$\pm$\textcolor{white}{0}8.59}                    & \text{11.44$\pm$\textcolor{white}{0}5.66}                    & \text{10.27$\pm$\textcolor{white}{0}5.64}                    & \text{\textbf{\text{\textcolor{white}{0}9.66$\pm$\textcolor{white}{0}5.57}}}  \\
 2\_50\_50    & \text{25.80$\pm$12.43}                                      & \text{24.67$\pm$15.99}                                                       & \text{20.65$\pm$16.46}                                       & \text{15.61$\pm$11.50}                                       & \text{14.64$\pm$\textcolor{white}{0}8.28}                    & \text{15.52$\pm$12.51}                                       & \text{\textbf{\text{11.57$\pm$\textcolor{white}{0}5.99}}}                     \\
\hline
\end{tabular}
\captionof{table}{MAPE, sewage data set, 6 tanks}
 \label{table:mapeResultTable}\vspace{-10pt}
\end{figure*}
%
\begin{figure*}[!h]
\begin{center}
\begin{tabular}{rccccccc}
\hline
|X|, l, p   & YESTERDAY   & ARIMA         & RNN                        & LSTM       & MTCNN                       & CRNN   & AECRNN       \\
\hline
2\_50\_25    & \text{1.02} & \text{1.04} & \text{0.66} & \text{0.64}                 & \text{0.60} & \text{\textbf{\text{0.53}}} & \textbf{0.53} \\
 2\_50\_50    & \text{1.10} & \text{1.18} & \text{0.67} & \text{\textbf{\text{0.60}}} & \text{0.73} & \textbf{0.60}                 & \text{0.62} \\
\hline
\end{tabular}
\end{center}
\captionof{table}{RMSE, Google Trends data set.
}
 \label{table:rmseResultTrends}
\end{figure*}
\begin{figure*}[!ht]
\begin{tabular}{llllllll}
\hline
 |X|, l, p  & YESTERDAY     & ARIMA                           & RNN                          & LSTM         & MTCNN         & CRNN     & AECRNN         \\
\hline
2\_50\_25    & \text{392.66} & \text{700.16} & \text{\textbf{\text{337.09}}} & \text{403.97}                 & \text{469.04} & \text{347.68} & \text{395.33} \\
 2\_50\_50    & \text{407.11} & \text{671.68} & \text{368.47}                 & \text{\textbf{\text{363.02}}} & \text{462.82} & \text{425.61} & \text{484.09} \\
\hline
\end{tabular}
\captionof{table}{MAPE, Google Trends data set}
 \label{table:trendsMapeResult}
\end{figure*}

For the sewage data set, we started by using one tank (i.e., tank 4) and consider all three time series in the tank (see Tables~\ref{table:rmseResult} and ~\ref{table:mapeResult}). 
Afterwards, we decided to focus on 2 highly correlated time series---$NH4$ and $NO3$ and conduct experiments on all 6 tanks (see Tables~\ref{table:rmseResultTable} and ~\ref{table:mapeResultTable}).  

\begin{figure*}[h]
    \centering
    \captionsetup{justification=centering}
    \includegraphics[width=.49\linewidth, height=0.25\textheight]{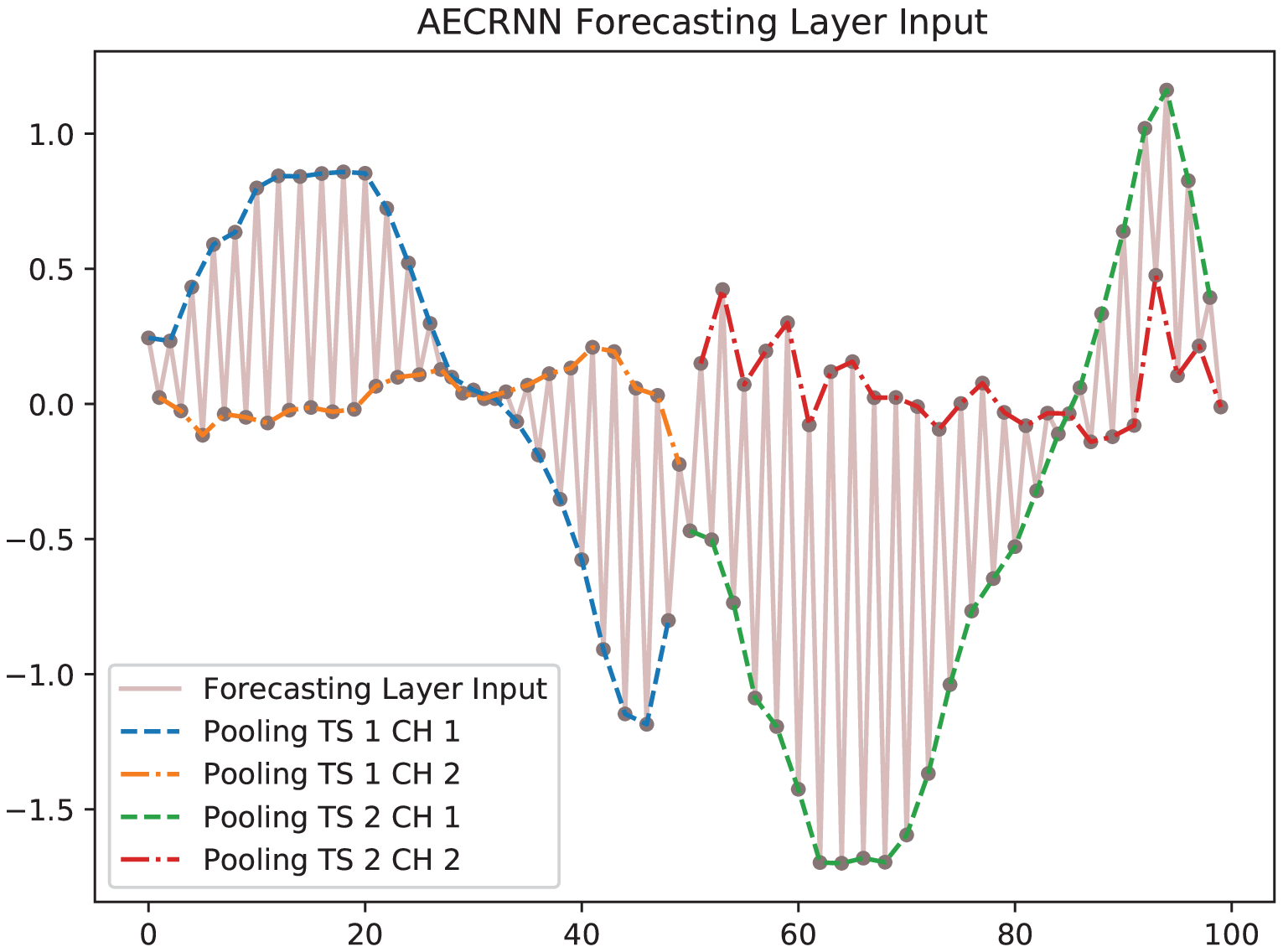}
      \includegraphics[width=.49\linewidth, height=0.25\textheight]{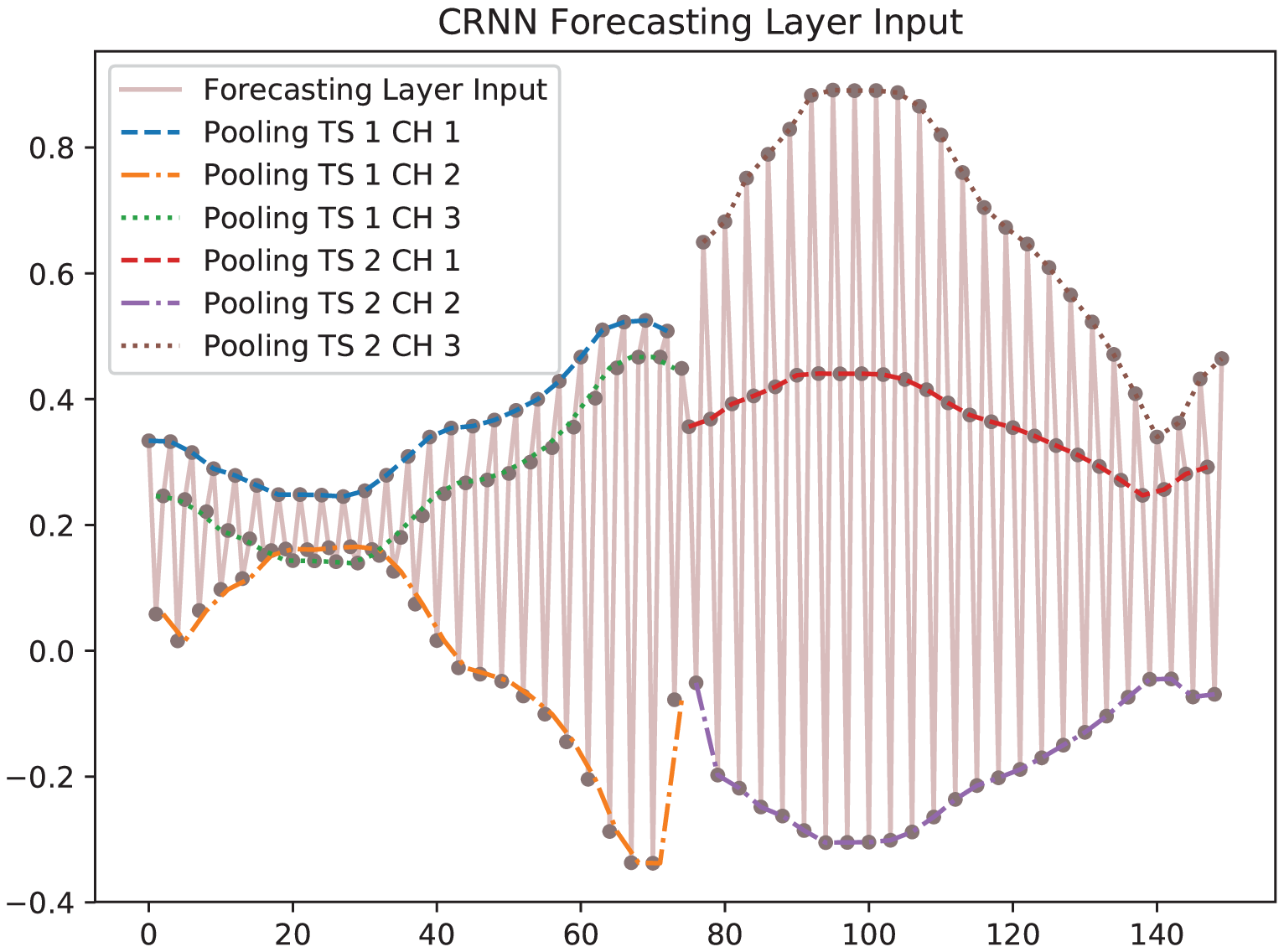}
    \caption{Input for the forecast layer of AECRNN (left) and CRNN (right), configuration 2\_50\_25, tank 4, sewage data set}
    \label{fig:forecasting_Input_example}
\end{figure*}
\begin{figure*}[h]
    \centering
    \captionsetup{justification=centering}
    \includegraphics[width=.49\linewidth, height=0.25\textheight]{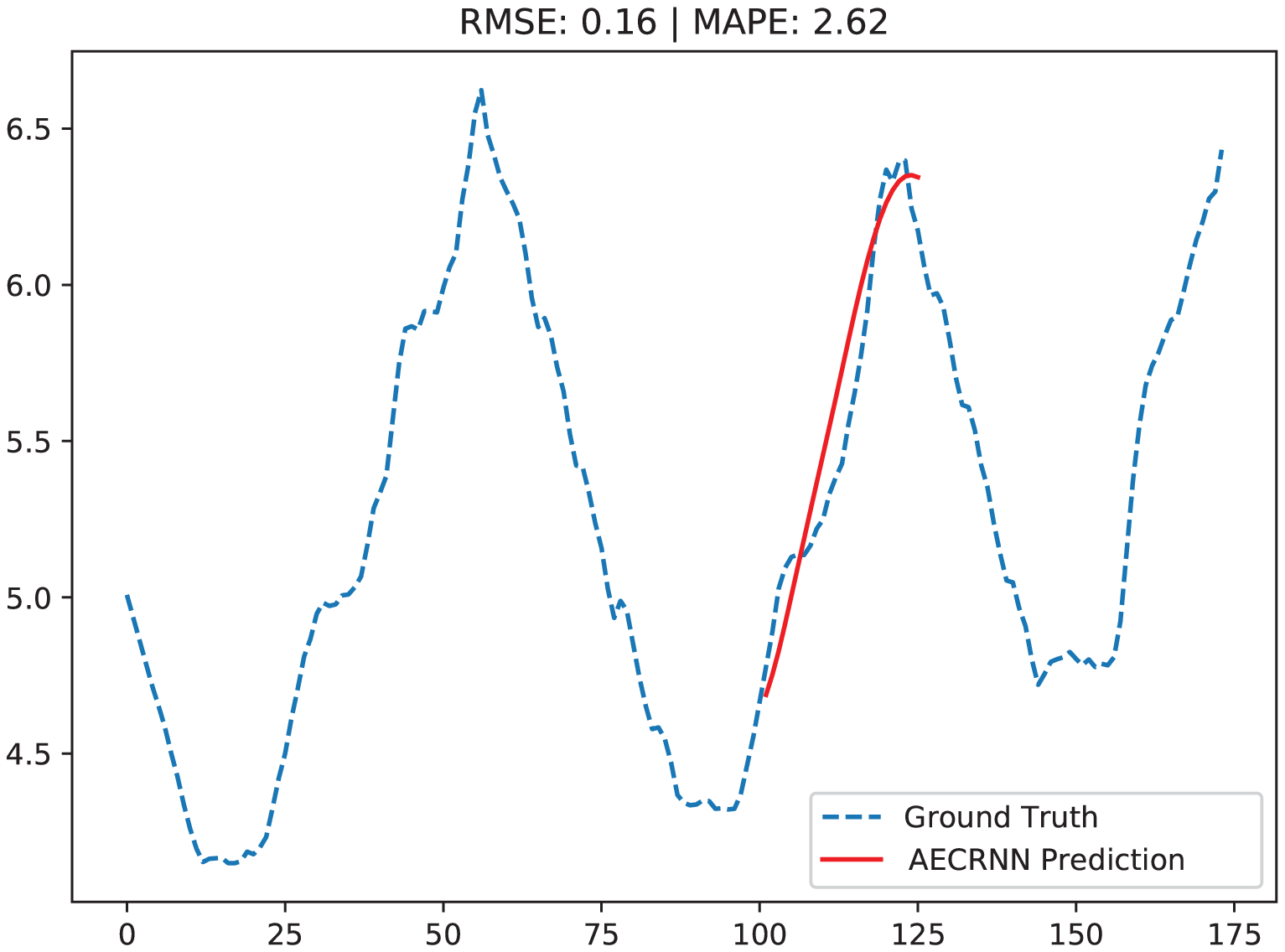}
    \includegraphics[width=.49\linewidth, height=0.25\textheight]{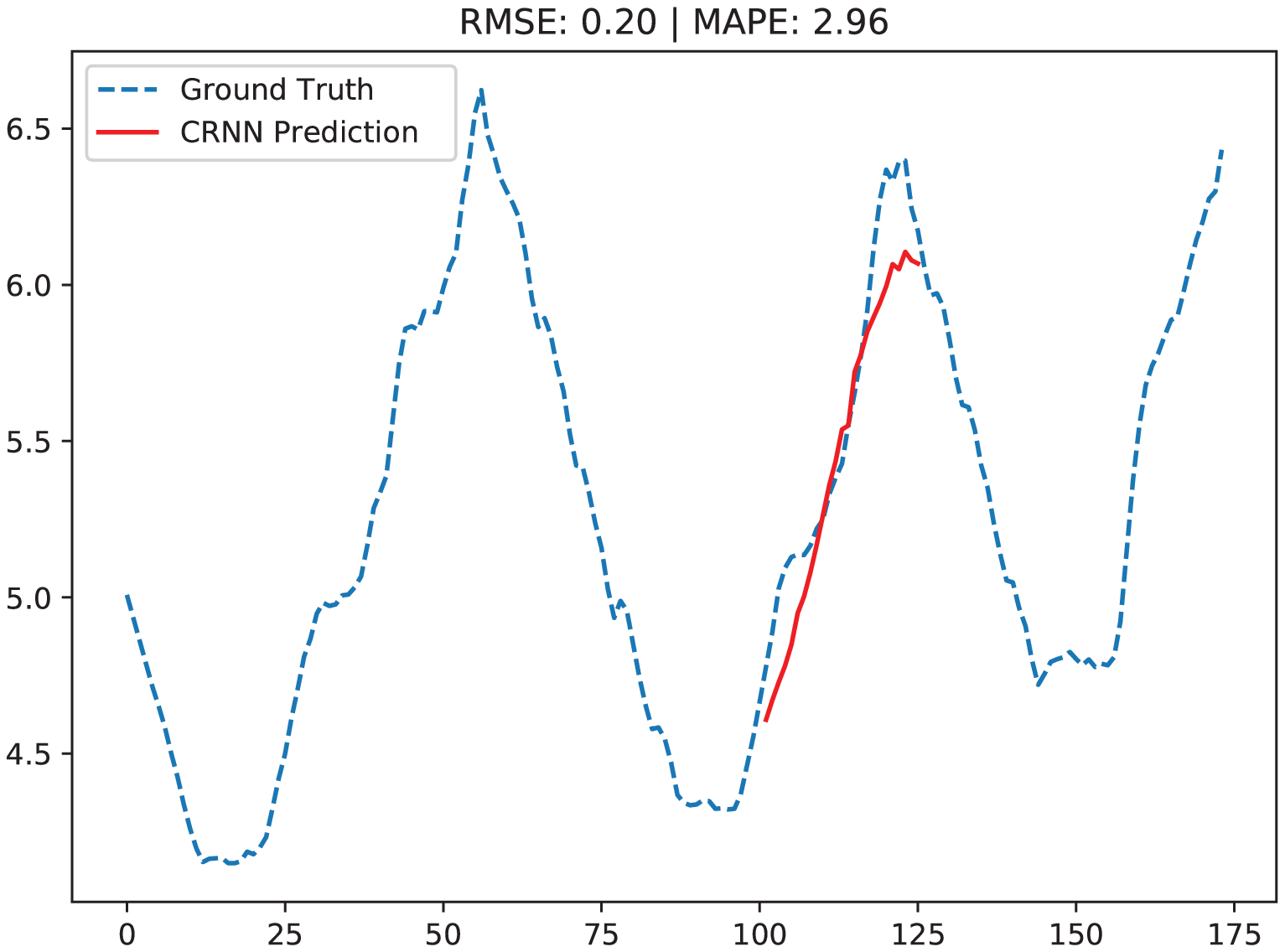}
    \caption{Prediction, AECRNN (left) and CRNN (right), configuration 2\_50\_25, tank 4, sewage data set}
    \label{fig:prediction_example_MAECNNRNN_MCRCNN}
\end{figure*}
\begin{figure}[h]
    \centering
    \captionsetup{justification=centering}
    \includegraphics[width=\columnwidth]{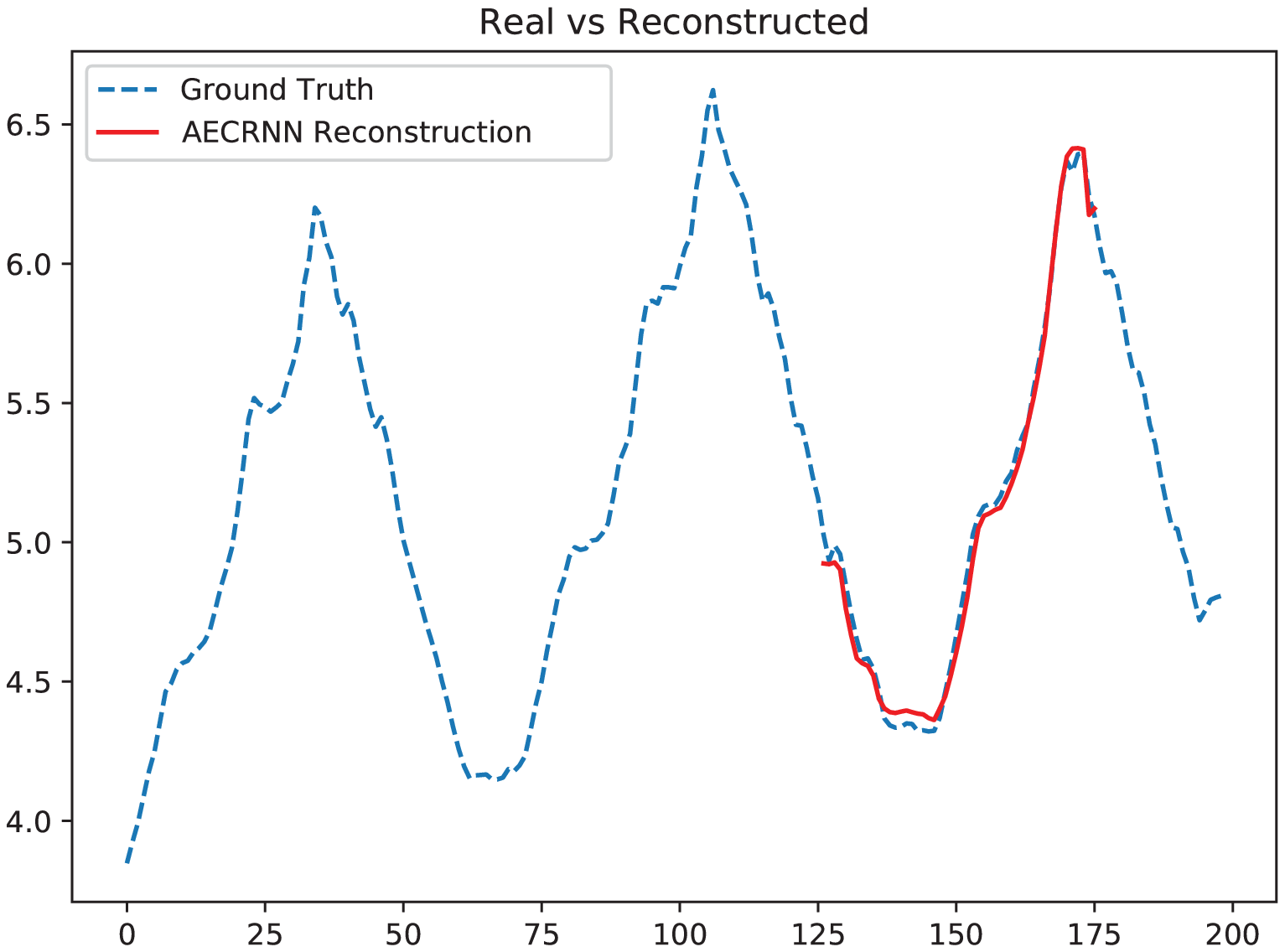}
    \caption{Reconstruction for AECRNN, configuration 2\_50\_25, tank 4, sewage data set}
    \label{fig:reconstruction_example_MAECNNRNN}
\end{figure}

When $|X|=1$, i.e., forecasting a single target time series, RNN has the best accuracy. However, when $|X|>1$, i.e., forecasting the target time series with multiple correlated time series, the proposed CRNN and AECRNN achieve the best accuracy. Note that MTCNN, CRNN, and AECRNN achieve better accuracy when considering correlated times series, meaning that they take advantage of correlated time series. In contrast, Yesterday and ARIMA cannot consider the additional correlated time series, and RNN and LSTM do not show clear improvements.  
When the input training sequence is very short, e.g., $l=10$, LSTM gives the best accuracy. As $l$ increases, CRNN and AECRNN achieves better accuracy compared to other methods. This suggests that CRNN and AECRNN are able to take advantages of having longer training sequences. 

When $p=1$, {ARIMA} gives very accurate prediction, which suggests that linear model is very good at short-term forecasting. However, when predicting far into the future , i.e., when $p$ is large, {ARIMA} deteriorates quickly and CRNN and AECRNN give more accurate forecasting.

Fig. \ref{fig:forecasting_Input_example} shows the input of the forecast layer of the proposed two models. For AECRNN we use 2 filters, and for CRNN we use 3 filers. As mentioned previously this hyper parameters were chosen for optimal results. In the figure, ``Pooling TS$i$ CH$j$'' stands for the output of the pooling layer of the time series $X^{(i)}$ from channel $j$. For AECRNN each filter yields a more human readable result, with one of the filters focused on a general representation of the input while the other tries to model the residuals. Without the auto-encoding layer, CRNN is granted more liberty in selecting whatever features it finds more usable, resulting in less human readable input.

Fig. \ref{fig:prediction_example_MAECNNRNN_MCRCNN} shows a concrete example of predictions made by the proposed two models when $n=2$, $l=50$, and $p=25$. It can be observed that AECRNN is able to predict better the amplitude of the spike, but ignores the small distortions at the top. In contrast, MCRNN predicts a lower amplitude but does capture the distortions at the top. 
We consider this to be a consequence of the auto-encoding process that values the correct amplitude more. This behavior can be observed in Fig.  \ref{fig:reconstruction_example_MAECNNRNN} where red curve represents the reconstructed input generated by the auto-encoder.

%

\subsubsection{Results on Google Trend Data}
%
For the Google trends data set, we use different problem parameters. In particular, we keep $|X|=2$ and $l=50$ and only vary $p$ from 25 to 50. The reason is due to the limited size of the data.  

Table~\ref{table:rmseResultTrends} suggests that, when predicting $p=25$ steps ahead, the two proposed models outperform their competitors in terms of RMSE values. When $p=50$, AECRNN falls behind CRNN and LSTM which have similar results.  

The MAPE values shown in Table~\ref{table:rmseResultTrends} are high due to the fact that the Google Trend time series generally have very small values.

\subsubsection{Effect of Uncorrelated Time Series}
Finally, we conducted an experiment to show that AECRNN is able to provide robust forecasting when the input time series (TS) are uncorrelated. In particular, we consider three cases: a single target TS alone, the target TS with a correlated TS, and the target TS with a generated, uncorrelated TS. 
The results in Table~\ref{uncorr} suggest that (1) AECRNN is more robust to deal with the uncorrelated time series, due to the auto-encoders in AECRNN; (2) CRNN is able to take more advantage when having correlated time series.  

\begin{table}[!htb]
\centering \small
\begin{tabular}{lll}
\hline
    & CRNN        & AECRNN    \\
\hline
 Single target TS     & \text{10.8} & \text{10.7} \\
 With a correlated TS   & \text{8.3}  & \text{9.1}  \\
 With an uncorrelated TS      & \text{13.0} & \text{10.7} \\
\hline
\end{tabular}
    \captionof{table}{Dealing with uncorrelated time series, MAPE}
    \label{uncorr}
\end{table}
\section{Conclusion and Outlook}
We propose two deep learning models, CRNN and AECRNN, to enable accurate correlated time series forecasting. Experiments on two real world data sets show promising results. In the future, it is of interest to verify the proposed models on time series from different domains such as transportation~\cite{DBLP:journals/tits/Ding0GL15,DBLP:journals/tc/Ding0C016,l2r} and to study the scalability of the proposed models for large time series data, e.g., by exploring parallel computing frameworks~\cite{DBLP:conf/dasfaa/YangMQZ09,DBLP:conf/waim/YuanSWYZY10}.
\label{conclusion}

\end{document}